\documentclass[letterpaper, 10 pt, conference]{ieeeconf}  

\IEEEoverridecommandlockouts                              

\pdfoutput=1
\usepackage{graphics} 
\usepackage{epsfig} 
\usepackage{mathptmx} 
\usepackage{times} 
\usepackage{amsmath} 
\usepackage{amssymb}  
\usepackage{graphicx}
\usepackage{bbding}
\usepackage{balance}

\title{\LARGE \bf
A Robust Quadruped Robot with Twisting Waist for Flexible Motions
}

\author{Quancheng Qian$^{1}$, Xiaoyi Wei$^{1}$, Zonghao Zhang$^{1}$, Jiaxin Tu$^{1}$, Yueqi Zhang$^{1}$,   \\ Taixian Hou$^{1}$, Xiaofei Gao$^{2}$, Peng Zhai$^{1,*}$ and Lihua Zhang$^{3,*}$
\thanks{$^{1}$Quancheng Qian, Xiaoyi Wei, Zonghao Zhang, Jiaxin Tu, Yueqi Zhang,  Taixian Hou, Peng Zhai, are with the Academy for Engineering and Technology, Fudan University, Shanghai 200433, China
        {\tt\small qcqian24@m.fudan.edu.cn; pzhai@fudan.edu.cn}}%
\thanks{$^{2}$Xiaofei Gao is with Beijing Jingcheng Zhitong Robotics Technology Company, Beijing, China {\tt\small gaoxiaofei@inter-smart.com}}%
\thanks{$^{3}$Lihua Zhang is with the Engineering Research Center of AI and Robotics, Shanghai, China
        {\tt\small lihuazhang@fudan.edu.cn}}%
\thanks{*Corresponding Author.}
}

\begin{document}
\maketitle
\thispagestyle{empty}
\pagestyle{empty}

\begin{abstract}

The waist plays a crucial role in the agile movement of many animals in nature. It provides the torso with additional degrees of freedom and flexibility, inspiring researchers to incorporate this biological feature into robotic structures to enhance robot locomotion. This paper presents a cost-effective and low-complexity waist mechanism integrated into the structure of the open-source robot solo8, adding a new degree of freedom (DOF) to its torso. We refer to this novel robot as solo9. Additionally, we propose a whole-body control method for the waist-equipped quadruped robot based on generative adversarial imitation learning (GAIL). During training, the discriminator is used as input for iterative optimization of the policy and dataset, enabling solo9 to achieve flexible steering maneuvers across various gaits. Extensive tests of solo9’s steering capabilities, terrain adaptability, and robustness are conducted in both simulation and real-world scenarios, with detailed comparisons to solo8 and solo12, demonstrating the effectiveness of the control algorithm and the advantages of the waist mechanism.


\end{abstract}

\section{INTRODUCTION}

In nature, the agility and flexibility of many animals are fundamentally supported by the waist as a core source of power. For instance, cheetahs rely on their waist strength to maintain stability during high-speed chases \cite{c1}. Similarly, cats can rotate their waist in mid-air to ensure a stable four-legged landing \cite{kane1969dynamical}. Inspired by these natural mechanisms, recent research has focused on improving and optimizing the torso structure of quadruped robots to achieve more agile locomotion \cite{xiang2024modeling}, \cite{khoramshahi2013benefits}, \cite{lin2023exploring}, \cite{huang2024optimizing}. For example, \cite{tang2023towards} enhanced the agility and landing safety of robots by adding an inertial tail with three DOFs to the robot's torso. \cite{wang2024spined} introduced a four DOFs spine to enable the robot to follow various gaits with high speed. While these efforts have advanced the locomotion capabilities of quadruped robots in different aspects, it is important to acknowledge that incorporating complex multi-degree-of-freedom mechanisms often presents significant control challenges.

\begin{figure}[htbp]
\centerline{\includegraphics[width=8.4cm]{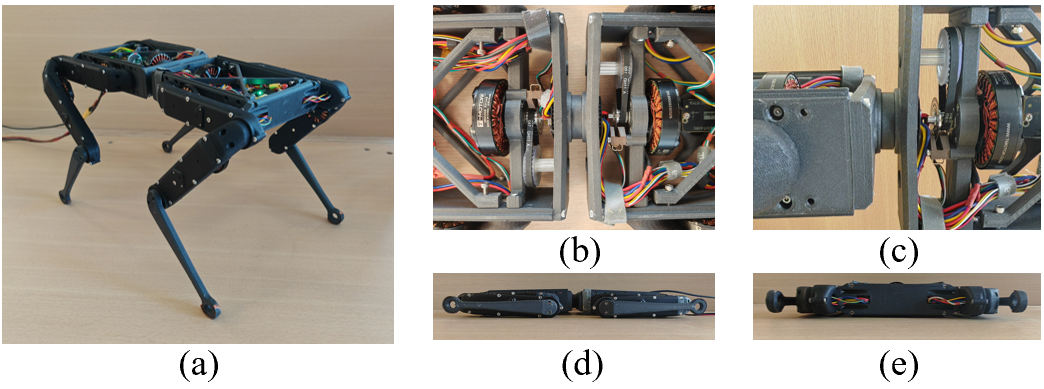}}
\vspace{-0.2cm}
\caption{(a) The 9-DOF quadruped robot ‘solo9’. (b) Close-up of the waist mechanism. (c) Close-up of the waist twisted 90 degrees. (d) Front views of solo9 after folding. (e) Side views of solo9 after folding.}
\label{fig}
\vspace{-0.4cm}
\end{figure}

To address these challenges, some research has explored simpler mechanisms to enhance locomotion performance by incorporating low-degree-of-freedom spines and employing a comprehensive control framework to better integrate with existing quadruped robots. However, current research on low-degree-of-freedom spines has predominantly focused on radial contraction and expansion \cite{bhattacharya2019learning}, \cite{zhao2017effect}, applied to tasks such as high-speed movement and climbing. In contrast, studies on the axial rotation capabilities of spines are mainly focused on simulations \cite{caporale2023twisting}. Intuitively, axial rotation of the spine could improve the robot's stability during turning and enhance maneuverability. Nonetheless, current mainstream robots primarily rely on fine control of hip joints, such as adduction and abduction movements, to adjust direction \cite{aractingi2023controlling}. There is limited research on utilizing spinal axial rotation to address agile turning challenges.



In this paper, we propose a novel torque-controlled quadruped robot system, named solo9, based on the open-source quadruped robot solo8 \cite{grimminger2020open}. The solo9 features a rotatable waist mechanism equipped with two high-torque brushless motors and a low-ratio gearbox. The robot's base is divided into two symmetric halves, with each half housing a rotational motor near the waist. This configuration not only endows solo9 with flexible steering capabilities not present in solo8 but also enhances the robot's robustness in navigating complex terrains and handling disturbances. Additionally, we introduce a dataset-policy co-optimization algorithm for training the solo9 controller, which leverages the motion dataset from solo8  to efficiently train solo9 across various gaits. Extensive testing has been conducted in both simulation and real-world scenarios to validate the advantages brought by the waist mechanism. Our contributions include:


\begin{itemize}
\item A novel, low-complexity rotatable waist mechanism for the quadruped robot solo9. 
\item A whole-body control method for solo9, based on generative adversarial imitation learning (GAIL), which includes transfer optimization of the policy and dataset to achieve flexible steering across various gaits.
\item Extensive testing of solo9's steering capabilities, terrain adaptability, and robustness in both simulation and real-world scenarios, with detailed comparisons to solo8 and solo12, demonstrating the effectiveness of our control algorithm and the rationality of the mechanical structure.
\end{itemize}

\section{RELATED WORK}

\subsection{Application of Spine and Waist Structure in Legged Robots}

In recent years, inspired by biological systems and advancements in control algorithms, there has been a growing presence of biomimetic robots. Both the spine and waist, as core components for controlling animal locomotion, have attracted significant research attention. Numerous quadruped robots now incorporate spinal structures \cite{xiang2024modeling}, \cite{khoramshahi2013benefits}, including models such as INU \cite{duperret2017empirical} and MIT Cheetah \cite{bledt2018cheetah}. Research on the spine and waist encompasses its impact on energy efficiency during walking \cite{caporale2023twisting}, its role in adjusting the robot's aerial posture \cite{zhao2017effect}, \cite{nakano2012control}, \cite{roscia2023orientation}, and its potential to enhance locomotion performance \cite{lin2023exploring}, \cite{bhattacharya2019learning}. The work that is closest to ours is \cite{caporale2023twisting}, but it mainly focuses on the influence of axial motion of the waist on the robot's motion loss and the realization of parkour-style wall-jumping, and it is only attempted in a simulation environment without extensive exploration of multi-gait, continuous turning tasks, or real robots.

\subsection{Reinforcement Learning Control for Quadruped Robots}



In recent years, the research on motion control algorithms for quadrupedal robots based on reinforcement learning has made significant progress, enabling the robots to perform various difficult and impressive actions and tasks \cite{hwangbo2019learning},  \cite{lee2020learning}, \cite{miki2022learning}. However, traditional reinforcement learning methods require the design of complex reward functions, and the weighting parameters of each reward function also need to be carefully tuned. 

An alternative approach for learning robot strategies is imitation learning, which simply compares actions with a reference dataset without the need for complex reward functions \cite{peng2021amp}. Nevertheless, relying solely on imitation learning algorithms also suffers from issues such as poor policy transferability and high quality requirements for imitation datasets. To address these issues, many recent studies have incorporated both task rewards and imitation rewards into the reward system \cite{peng2020learning}, \cite{peng2018deepmimic}. Some recent studies have used the solo8 quadrupedal robot to optimize and fine-tune policies based on task rewards while imitating reference datasets \cite{li2023learning}, \cite{li2023versatile}, achieving stable walking in various gaits.

\begin{figure}[htbp]
\centerline{\includegraphics[width=9cm]{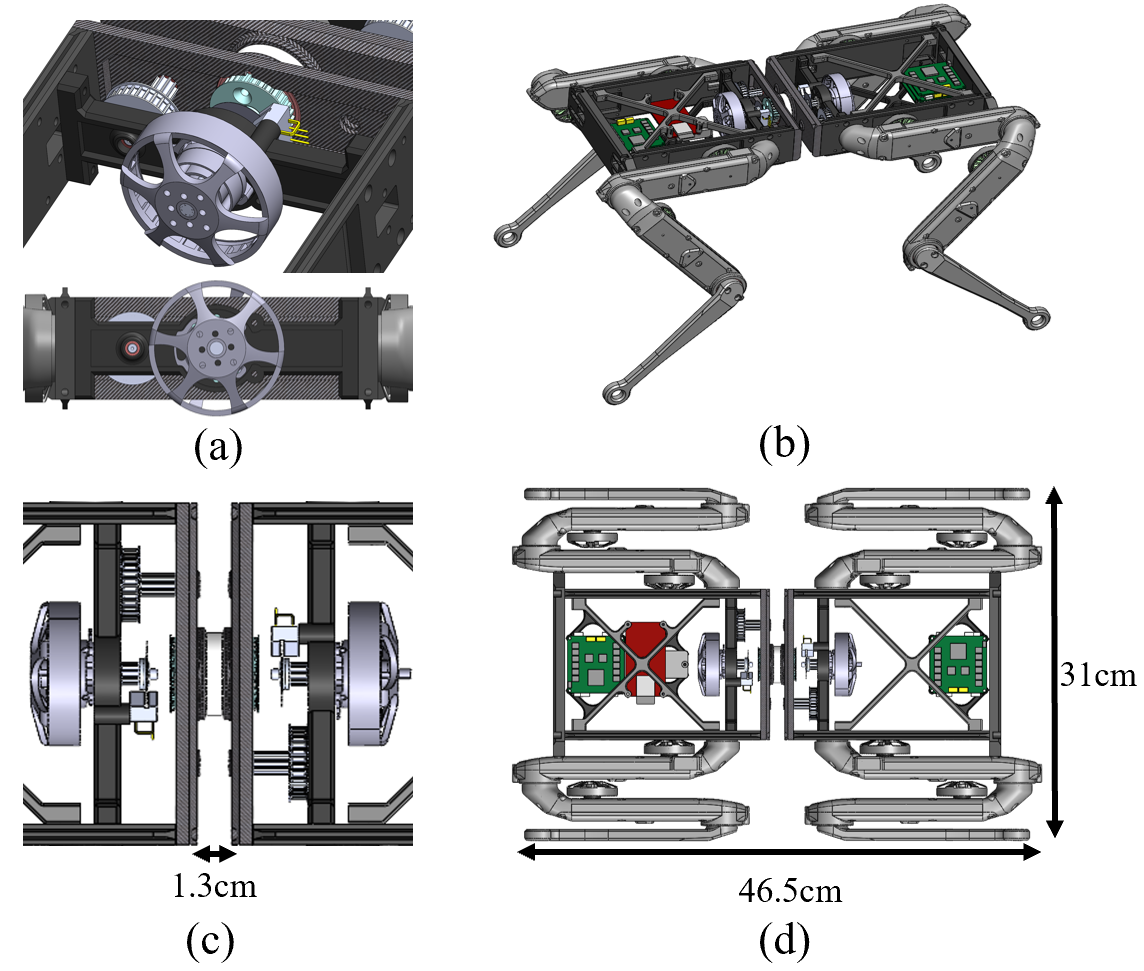}}
\vspace{-0.2cm}
\caption{Schematic diagram of solo9's mechanical structure: (a) (c) show the waist motor structure, (b) is the whole-body view, (d) is the top-down view.}
\vspace{-0.3cm}
\label{fig}
\vspace{-0.3cm}
\end{figure}


\subsection{Whole-body Control of Legged Robots}


The introduction of additional mechanical structures onto quadrupedal robots further increases the complexity of system control. Huang et al. \cite{huang2024manipulator} introduced a robotic arm onto a quadrupedal robot and achieved coordinated control through a multi-stage hierarchical training approach. However, recent studies have shown that using hierarchical models in a semi-coupled manner to control leg and arm movements is often ineffective as it can easily fall into local optima \cite{fu2023deep}. Whole-body control is an elegant and concise control concept that treats all parts of the controlled object as a whole, attempting to consider problems within a unified framework to avoid falling into local optima \cite{bellicoso2017dynamic}. This concept has been widely applied in legged robots, especially humanoid robots, has achieved good results \cite{sentis2006whole}. In this paper, we adopt the whole-body control concept to generate a unified controller for the waist and legs of the solo9 robot.

\section{METHOD}


\subsection{Overview of the Solo9 Robot}
\subsubsection{Mechanical Structure} The solo9 builds upon the solo8 by modifying the base into an axially rotatable front-rear symmetric structure. Rotation is achieved through drive motors (T-MOTOR MN5008 340KV) located near the waist on each side, where the two motors rotate in opposite directions. Field-Oriented Control (FOC) and high-precision encoders are employed to ensure that the rotational speed of each motor remains consistent. Each motor is connected to a 9:1 bipolar belt transmission, with a hollow central axis used to pass through power and communication cables. Mechanically, we do not impose any limitations on the rotation angle of the robot's waist. The solo9 robot has a length of 46.5cm, a width of 31cm, and weighs approximately 2.3kg. In comparison, the replicated solo8 robot that we have created measures 42.8cm in length, 30.9cm in width, and weighs approximately 1.9kg.

\subsubsection{Electronics} The solo9 uses the same brushless motor driver boards as the solo8 for Field Oriented Control (FOC) of the motors. Additionally, we have added an extra control board to drive the two motors in the waist. The master board controls five FOC driver boards, with three placed on the rear base to drive the rear legs and waist motors on both sides, and two placed on the front base to drive the front legs on both sides. The master board and IMU are mounted at the front end, and the master board can connect to the control computer via Ethernet or WiFi. The robot uses the IMU to sense the pose of the front end of the body and calculates the pose of the rear end of the body by combining the waist motor angles with the IMU data.

\subsubsection{Simulation Model}We have created a URDF file for the solo9, with the front base as the root link, connecting the two front legs to the rear base, which is then connected to the two rear legs. Since Isaac Gym does not support a single joint being driven by multiple motors, we use only a single motor to drive the waist rotation in the simulation environment. As the actual machine uses two motors to drive the waist joint, the motor output torque in the simulation model is twice that of a single motor in the real machine. To represent the angular position between the front and rear bases, we define joint angles between the bases. Similar to the leg joints, we include the joint angles and angular velocities between the front and rear bases in the observations.



\subsection{Whole-body Control Algorithm for Solo9}

The solo9 robot, with the addition of an x-axis articulated waist mechanism compared to solo8, introduces complexities in its control system. Numerous intricate reward functions need to be considered to maintain the stability of the waist structure and its synergy with the original robot's architecture. Fortunately, the research on motion control for the solo8 robot is now relatively mature, with various high-quality gait datasets being open-sourced \cite{li2023versatile}. Considering the complexity of designing reward functions directly through reinforcement learning methods and the structural similarity between solo9 and solo8, we employ a reinforcement learning-based imitation learning approach to achieve whole-body control of solo9.


\subsubsection{RL Basic Architecture} The action space of solo9 is defined by the torques of 9 motors, including one waist motor and 8 leg motors, with 2 motors driving each thigh and calf. The basic state space is constructed based on the robot's proprioception and consists of 33 dimensions: the base quaternion (4 dimensions), two control commands including forward speed and yaw rates, 9 joint positions (9 dimensions), speeds (9 dimensions), and the previous time step's positions (9 dimensions). This data is obtained from brushless motor driver boards and sensor readings from an Inertial Measurement Unit (IMU) in practice. Due to the difficulty in accurately measuring the base linear velocity in real-world environments, we use an asymmetric actor-critic structure. The actor network has an input dimension of 33-dimensional basic state space, while the critic network adds the base linear velocity as a privileged observation (which can be directly obtained in simulation environments), making the state space 36 dimensions.

\begin{figure}[htbp]
\centerline{\includegraphics[width=9cm]{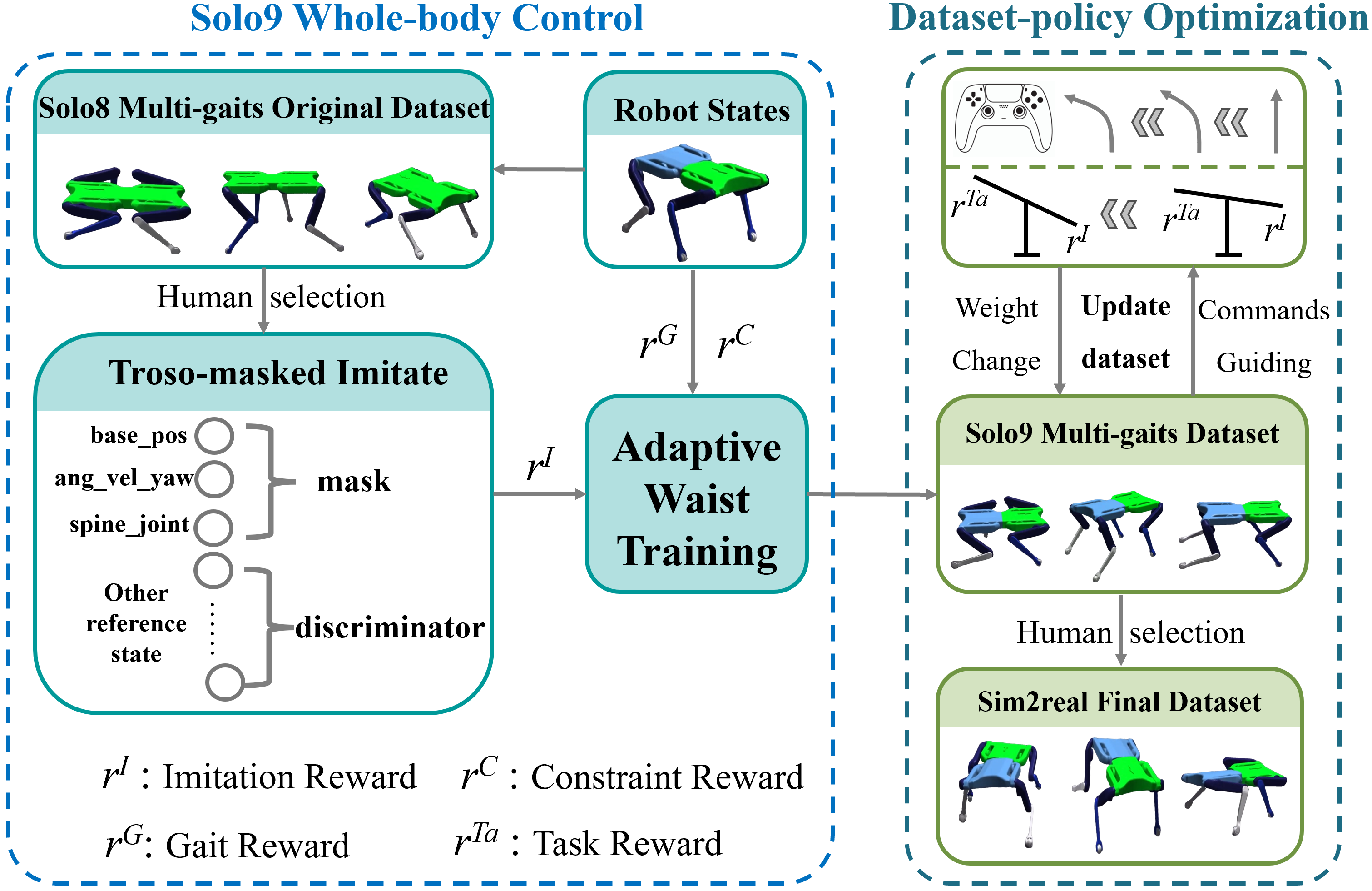}}
\caption{The figure illustrates the dataset-policy co-optimization control algorithm for solo9 based on GAIL. Starting with the solo8 original dataset, and after manual selection, the discriminator inputs partial observations and combines gait and constraint rewards for adaptive training of the waist, resulting in the initial solo9 dataset. Further dataset-policy co-optimization with command guidance is performed, increasing the weight of imitation rewards through iterations, ultimately producing the solo9 flexible steering dataset.}
\vspace{-0.3cm}
\label{fig}
\end{figure}

\subsubsection{Basic Architecture of Imitation Learning} We use a Generative Adversarial Imitation Learning method \cite{ho2016generative}, considering the imitation observation space \( O^I \). The complete state space \( S \) of the underlying Markov Decision Process is mapped to the imitation observation space \( O^I \) through a function \( f \). We represent the state corresponding to a trajectory segment of length \( H_I \) up to time \( t \) as \( s_t = (s_{t-H_I+1}, \ldots, s_t) \). The discriminator is optimized using the Least Squares GAN (LSGAN) loss, with \( H_I \)-step inputs and gradient penalty. 


The observations input to the discriminator can include the following components: base position (3 dimensions), base quaternion (4 dimensions), base linear velocity (3 dimensions), base angular velocity (3 dimensions), base normal vector (3 dimensions), base height (1 dimension), joint positions (9 dimensions), and joint velocities (9 dimensions). During the initial training phase, base position and base yaw rate inputs are omitted as they interfere with the robot's steering training. The base quaternion and normal vector have similar representational capabilities here, so we use the normal vector and ignore the quaternion in training. Consequently, the discriminator observes a total of 27 dimensions: base linear velocity, base angular velocity, base normal vector, base height, joint positions, and joint velocities.


\subsubsection{Algorithm Control Flow} Firstly, based on the Isaac Gym environment \cite{makoviychuk2021isaac}, we perform preliminary training using the dataset open-sourced by Li et al. \cite{li2023versatile} to obtain various robust gaits for solo8, such as trot and leap. We then store the selected gaits and augment the sequence with waist information. Here, we set all waist data to zero to serve as the original gait dataset for training solo9.

Discriminator Observations: Since the original dataset for solo9 lacks useful waist information, directly applying imitation learning will make it difficult to effectively utilize the waist. To achieve this, we do not feed the discriminator with information related to the waist or other turning movements, encouraging the robot to learn turning movements utilizing its waist. If all observations were directly input to the discriminator, it would lead to conflicts between imitation rewards and steering rewards in the task rewards, affecting the training effectiveness. During this phase, extra gait rewards are introduced for fine-tuning to promote sim-to-real transfer \cite{aractingi2023controlling}.

Dataset \& Gait Co-Optimization: Since imitation rewards largely regulate the robot's behavior, we only need to adjust the weights of a few additional reward terms and store the action sequences generated by the new policy as a new reference dataset for iterative training. The co-optimization of policy and dataset plays a significant role in the gait learning process for solo9. It allows for the decoupling of different reward functions, steadily improves the quality of the dataset gaits, and gradually unlocks the potential of the waist in aspects such as steering and robustness. 

Reward Engineering: During the training process, we introduced a yaw rate tracking reward in the z-axis direction to encourage the robot to use its waist for timely steering adjustments.

\begin{equation}\label{key2}
r_{angvel}=c_{angvel} e^{||V^{cmd}-V_{w_z}||^2},
\end{equation}

where \( V^{cmd} \) is the angular velocity command. \( V_{w_z} \) is the yaw rate of the robot, and \( c_{angvel} \) is the angular velocity penalty weight.

Similar to \cite{hou2024multi}, \cite{aractingi2023controlling}, we also included a foot-lifting reward and a foot slipping penalty to prevent frequent contact between the feet and the ground.

\begin{equation}\label{key2}
r_{slip}=c_{slip}\sum_{i = 1}^4 C_i ||\dot{p}_{xy}||^2,
\end{equation}

where \( C_{i} \) is a filtered indicator of whether foot \(i\) is in contact with the ground, \( \dot{p_{xy}} \) is the foot’s plane speed, and \( c_{slip} \) is the slip penalty weight.
 
\begin{equation}\label{key2}
r_{clear}=c_{clear}\sum_{i = 1}^4 ||p_{z,i}-p^{max}_{z}||^2||\dot{p}_{xy}||^2,
\end{equation}

where \(i\) denotes the foot number, \(p_{z,i}\) is the expected foot
 height, we set a constant foot height target \(p^{max}_{z}\), and \(\dot{p}_{xy}\) stands for the velocity of the foot \(i\) in the x, y direction, and \( c_{clear} \) is the clearance penalty weight.
 
Additionally, the robot's motion frequency has a significant impact on sim-to-real transfer, both excessively high and low motion frequencies can lead to failures in actual deployment. To address this, we added an action\_rate constraint reward term to regulate the motion frequency. 

\begin{equation}\label{key2}
r_{smooth}=c_{smooth}||{q}^{target}_{t}-{q}^{target}_{t-1}||^2,
\end{equation}

where \( {q}^{target}_{t} \) and \( {q}^{target}_{t-1} \) is the position of the joint at time \(t\) and \({t-1}\), and \( c_{smooth} \) is the smooth penalty weight.

To facilitate analysis, we divide the total rewards \(r^{To}\) into imitation rewards \(r^{I}\) and task rewards \(r^{Ta}\). The task rewards are further categorized into gait rewards \(r^{G}\), constraint rewards \(r^{C}\), and turning rewards \(r^{Tu}\). 

\begin{equation}\label{key2}
r^{Ta}=r^{C}+r^{G}+r^{Tu}
\end{equation}

The key reward terms used in the experiments are grouped and presented in the Figure 4.

\begin{figure}[htbp]
\vspace{-0.1cm}
\centerline{\includegraphics[width=7cm]{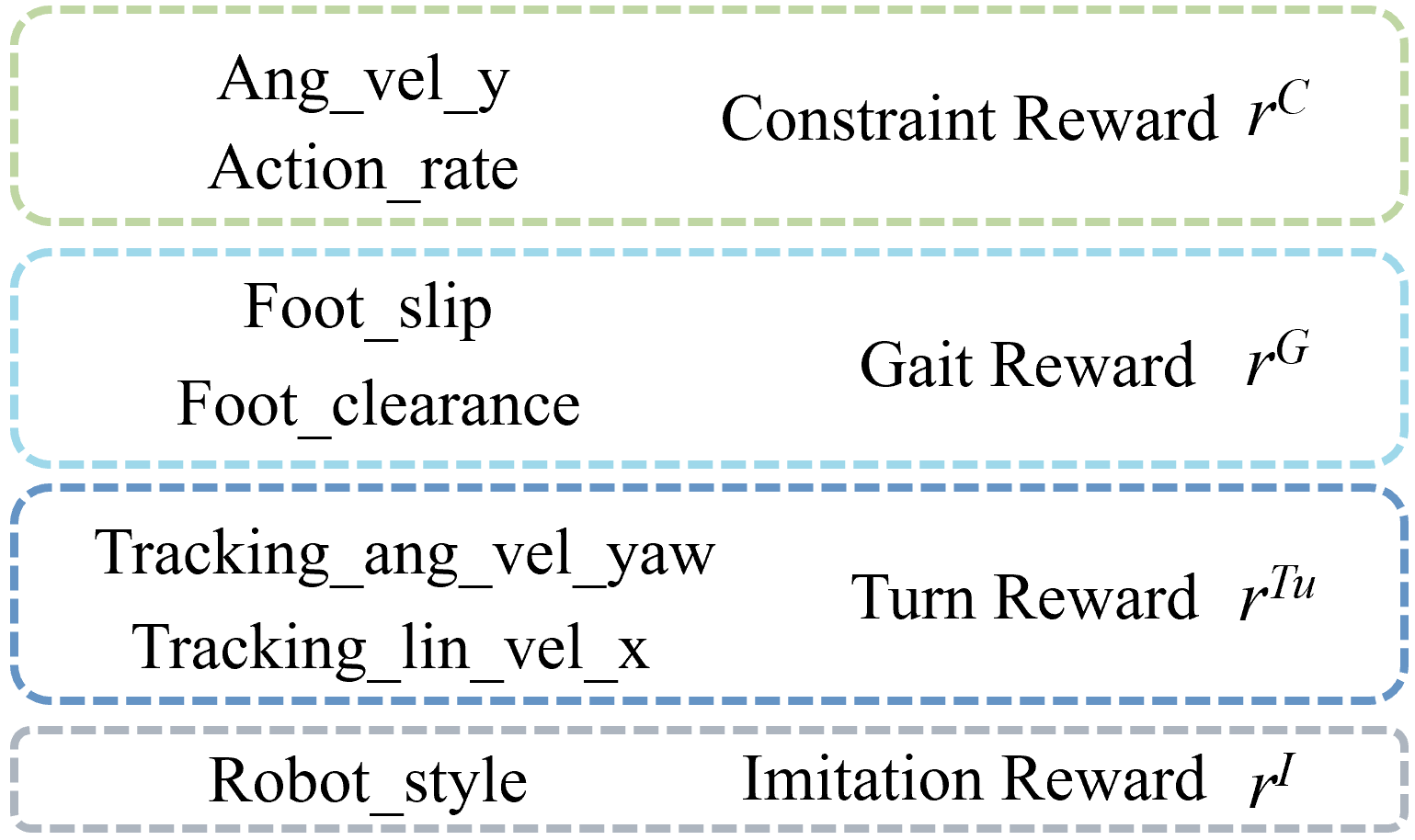}}
\caption{The key reward functions utilized in solo9's training}
\vspace{-0.3cm}
\label{fig}
\end{figure}

\subsection{Sim2real Transfer}


\subsubsection{Domain Randomization}To improve the robustness of the robot's policy, we performed extensive domain randomization during training. Building on the open-source legged\_gym project \cite{rudin2022learning}, we carried out extensive domain randomization on the robot's joint friction, base mass, initial joint angles, and center of mass \cite{margolis2023walk}. It is important to note that this work needs to be conducted after establishing the reference dataset, in order to minimize the impact of domain randomization on the solo9 gaits. TABLE I shows the details of domain randomization.

\begin{table}[htbp] 
\caption{DOMAIN RANDOMIZATION PARAMETERS}
\begin{center}
\begin{tabular}{@{}c|cccc@{}} 
\hline
Randomization              & Lower bound        & Upper bound                 \\ \hline
friction & 0.2          & 2.5                     \\           
base mass & -0.7          & 1.5                  \\  
center of mass & -1.5          & 1.5              \\
inital joint angles & 0.9          & 1.1                       \\
Linear velocity commands & 0          & 1                       \\
Angular velocity commands & -0.5          & 0.5                       \\
\hline
\end{tabular}
\end{center}
\vspace{-1.0em}
\end{table}


\subsubsection{Curriculum Learning} We used curriculum learning to train the PD parameters of the robot's motors and the height parameters of rugged terrain in the environment. These parameters significantly impact the complexity of the training. Using large parameter settings early in the training can easily cause the policy to fall into local optima, leading to decreased performance or even failure to learn effectively. Therefore, this paper adopts a curriculum learning approach to adjust these parameters, following the method from legged\_gym \cite{rudin2022learning}. If the robot achieves the required walking distance under the current parameters, it will then operate under more challenging conditions in the next reset. However, if the robot's movement distance is less than half of the target distance at the end of a period, the difficulty of the curriculum will be reduced. Robots that complete the highest level of the curriculum are cycled back to randomly selected levels to increase diversity and avoid catastrophic forgetting.


\begin{figure*}[htbp]
\centering
\includegraphics[width=17.6cm]{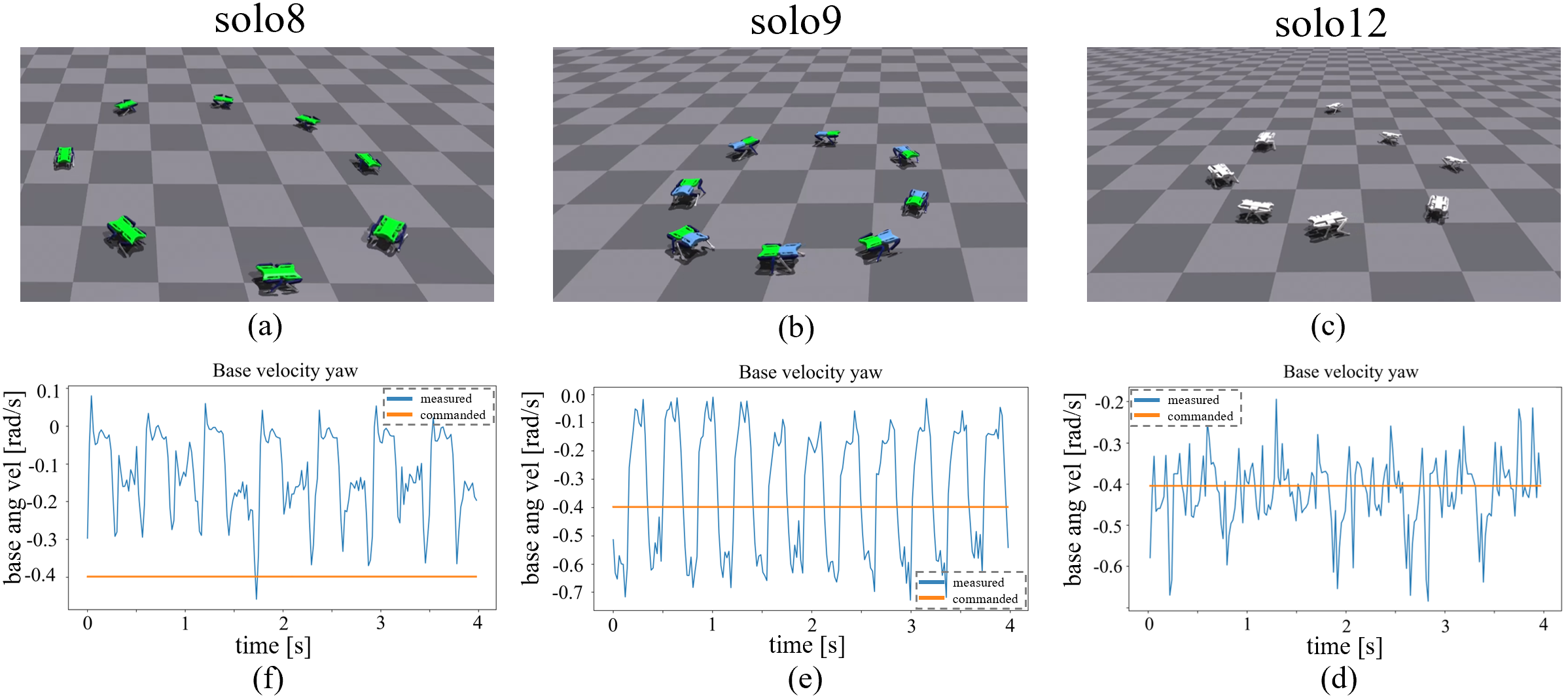}
\vspace{-0.2cm}
\caption{(a), (b), and (c) show the results of the turning task in the trot gait for solo8, solo9, and solo12 in Isaac Gym, respectively, with solo9 demonstrating the smallest and most stable turning radius. (d), (e), and (f) display the yaw rate tracking performance of solo8, solo9, and solo12 during the aforementioned turning task, respectively, with solo9 showing the most stable tracking performance.}
\vspace{-0.4cm}
\label{fig:res}
\end{figure*}
\section{EXPERIMENTS}

To demonstrate the reasonableness and effectiveness of the waist mechanism in solo9, we conducted tests on steering capability, terrain handling, and disturbance resistance in both simulation and real world. The experimental results show that, compared to solo8, although solo9 only adds one DOF, it possesses more flexible and stable steering capabilities, even outperforming the solo12 robot, which has a higher DOF. Additionally, effective waist control significantly improves the robot's motion robustness.

\subsection{Steering Capability Tests}
\subsubsection{Simulation Tests} We tested the turning capabilities of solo9 under three gaits: trot, leap, and crawl. For each gait, we recorded the commanded yaw rates and the actual yaw rate of the robot, while maintaining a forward linear velocity command of 0.6m/s. Tests were conducted on solo8, solo9, and solo12. As shown in Figure 5(d)(e)(f), when the yaw rates command was -0.4 rad/s, solo8's yaw rate tracking performance was significantly weaker than that of solo9 and solo12. Compared to solo12, solo9 exhibited more stable yaw rates tracking and a smaller turning radius. This highlights the advantages of waist steering over hip steering.


\subsubsection{Real-World Tests} We controlled solo9 to complete specified steering tasks, including 180-degree turns (Figure 6(f)), circular movements (Figure 6(h)), "S" turns(Figure 6(g)), and even performed in-place turning on a high-friction carpet using the waist (Figure 6(b)). Solo9 successfully completed all these tasks. In contrast, although solo8 demonstrated some steering tracking capability in simulation, the lack of hip adduction and abduction DOFs led to irregular and uncontrollable steering deviations during real-world deployment, causing a significant loss of yaw rate tracking capability. Due to hardware limitations, we did not test the steering capability of solo12 in the real world, but simulations indicate that solo9 possesses comparable or even superior steering abilities to solo12, with fewer joint DOFs.

\subsection{Terrain Tests}
\subsubsection{Simulation Tests}We compared the stability of solo8 and solo9 on complex terrains in both simulation and real environments to verify the adaptability of the waist mechanism to different terrains. In the simulation environment, we generated two types of rugged terrains for testing: irregular ground with a maximum and minimum height of ±3.5 cm, and irregular steps with initial heights randomized at 2.5 cm, 2.75 cm, and 2.9 cm. Both robots were trained on flat terrain.


\begin{figure*}[htbp]
\centering
\includegraphics[width=17.6cm]{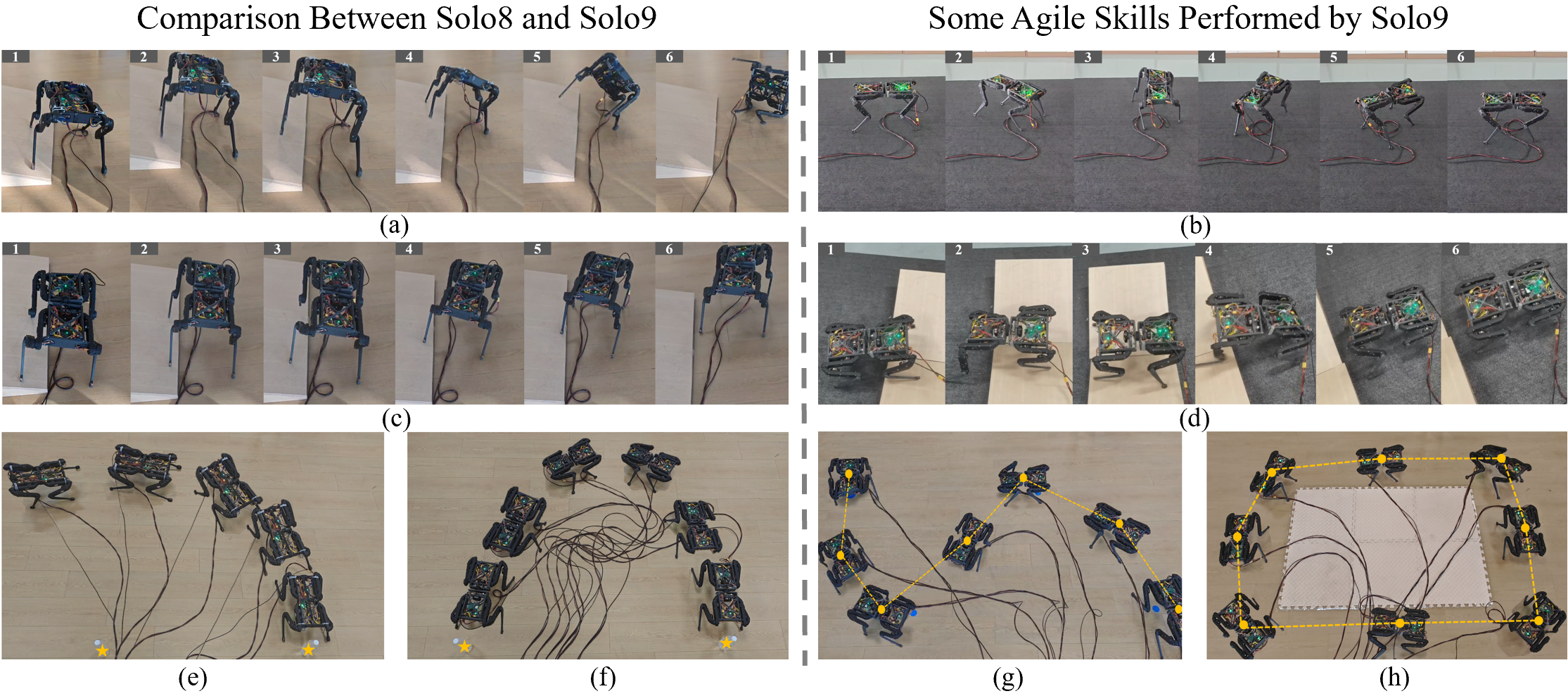}
\vspace{-0.4cm}
\caption{On the left, the real-world comparison experiments between solo8 and solo9 are shown. (a) and (c) display the performance of solo8 and solo9, respectively, when attempting to transition from an inclined slope to flat ground. (e) and (f) show the comparison of solo8 and solo9 in completing a 180-degree turn at a specified position.On the right, various agile skills of solo9 are demonstrated. (b) illustrates solo9 performing a 180-degree turn in place. (d) shows solo9 (trained only on flat ground) successfully navigating challenging terrain. (g) and (h) depict solo9 walking through an "S"-shaped curve (following the blue landmarks in the image) and around a square, respectively.}
\vspace{-0.5cm}
\label{fig:res}
\end{figure*}

We recorded the survival rate of the robots after running for 15 seconds to characterize their robustness on complex terrains. The experimental results, as shown in TABLE II, indicate that solo9 has a higher survival rate across all gait patterns on both uneven ground and irregular steps compared to solo8. Due to its lower base height, the crawl gait is prone to collisions with the ground, resulting in a higher survival rate compared to the other two gaits. The TABLE II demonstrates that the twisting of the waist has a significant effect on improving the robot's lateral stability.

\begin{table}[htbp] 
\caption{SURVIVAL RATES OF SOLO9 AND SOLO8 ACROSS VARIOUS GAITS ON IRREGULAR STEPS AND RUGGED GROUND.}
\begin{center}
\begin{tabular}{@{}c|c|ccc@{}} 
\hline
terrain & gait              & solo8        & solo9        & increase   \\ \hline
 &leap & 86.24±5.24\%          & 99.60±0.63\%             & 13.36\%          \\           
 uneven ground &crawl & 82.78±6.62\%          & 87.62±2.67\%          & 4.82\%            \\                        &trot & 88.58±7.31\%          & 99.00±0.54\%          & 10.42\%               \\ \hline   
  &leap & 81.14±3.62\%          & 92.25±1.25\%             & 11.11\%          \\           
 irregular steps &crawl & 58.94±6.78\%          & 71.51±3.23\%          & 12.57\%            \\                        &trot & 56.17±5.49\%          & 93.04±1.58\%          & 36.87\%               \\      
\hline
\end{tabular}
\end{center}
\vspace{-1.0em}
\end{table}

\subsubsection{Real-World Tests} We used the inclined ramp(with an angle of 25 degrees to the ground) shown in Figure 6(a)(c) to evaluate the robustness of solo8 and solo9 on complex terrain. We designed two sets of experiments: walking on an inclined surface as shown in Figure 6(a), and a terrain traversal task as shown in Figure 6(d). Both experimental scenarios involve unknown environments not encountered during simulation.

The results indicate that solo9 can dynamically adjust the relative position of the front and rear torso using its rotatable waist, allowing its legs to effectively support the ground and absorb stress, thus achieving greater stability on complex terrain. In contrast, due to its fixed torso, solo8 frequently experiences moments of leg missteps when traversing inclined surfaces, making it more prone to tipping over and potentially causing damage to the robot.

\subsection{Disturbance Resistance Tests}
Finally, we tested the adaptability of solo8 and solo9 to external disturbances in a simulation environment. Both robots were trained in the same environmental configuration. In the test environment, we applied random velocity disturbances with magnitudes of ±0.5m/s, ±0.7m/s, and ±1m/s in the x and y directions to the base of the robot at each time step, and we calculated the survival rates of the robot after operating for 15 seconds. TABLE III summarizes the survival rates for solo8 and solo9. The results show that as disturbances increase, the success rates of both robots decrease. However, thanks to the waist's adaptive response to disturbances, solo9 consistently outperforms solo8 in disturbance resistance, achieving up to a 15.6\% improvement in success rate under high disturbances.

Additionally, we conducted experiments on solo9's waist control. In the experiments, solo9 fixed refers to the waist being fixed at a 0-degree angle, while solo9 free means the waist is not actively controlled and allowed to move freely. The results indicate that when the waist is fixed, solo9 exhibits similar performance to solo8 in simulations due to their similar structures, resulting in comparable success rates. However, if the waist is not controlled and allowed to move freely, solo9 is unable to generate stable gaits. This underscores the necessity of a tailored motion control strategy to fully leverage the advantages of the waist mechanism.

\begin{table}[htbp]
\caption{DISTURBANCE RESISTANCE TEST RESULTS.}
\begin{center}
\begin{tabular}{@{}c|cccc@{}} 
\hline
robot              & push-0.5        & push-0.7        & push-1          \\ \hline
solo8 & 87.72±3.24\%          & 74.54±4.65\%             & 63.91±6.84\%          \\           

solo9 & 93.00±1.37\%          & 85.49±2.25\%          & 79.46±4.19\%         
              \\
solo9 fixed & 86.35±4.15\%          & 75.38±3.89\%          & 65.03±3.26\%            \\  
solo9 free & 0          & 0          & 0               \\
\hline
\end{tabular}
\end{center}
\vspace{-2.0em}
\end{table}


\section{CONCLUSION}
This paper presents the design of a novel quadrupedal robot, solo9, featuring a simple yet efficient waist that allows for axial rotation. Furthermore, we propose a whole-body control algorithm for quadrupedal robots with a waist. Leveraging datasets collected from the solo8 robot, we utilize the GAIL approach, coupled with command guidance and multiple rounds of dataset-policy iteration, to enable solo9 to achieve agile turning and robust locomotion across various gaits. We conducted extensive experiments in both simulation and real environments, and the results demonstrate that, despite having only one additional DOF compared to solo8, solo9 exhibits greater mobility, can track larger yaw rates, and even outperforms the solo12 robot which has a more DOFs. The survival rate of solo9 on rugged terrain and under external disturbances is significantly higher than that of solo8, indicating that the rotatable waist structure greatly aids the adaptability of legged robots to complex environments.

In future work, we plan to use solo9 to tackle more complex control tasks, such as aerial somersaults and yaw rates control in standing postures, which demand higher agility and precision. Additionally, we aim to enhance the waist mechanism by adding more DOFs, designing a robot that is more flexible, agile, and closely mimics the spinal structure of natural vertebrates.

\addtolength{\textheight}{-0cm}   









\bibliographystyle{IEEEtran}
\bibliography{mylib}

\end{document}